\documentclass[lettersize,journal]{IEEEtran}
\usepackage{amsmath,amsfonts}
\usepackage{algorithmicx}
\usepackage{algorithm}
\usepackage{algpseudocode}

\usepackage{array}
\usepackage[caption=false,font=normalsize,labelfont=sf,textfont=sf]{subfig}
\usepackage{textcomp}
\usepackage{stfloats}
\usepackage{url}
\usepackage{verbatim}
\usepackage{graphicx}
\usepackage{cite}
\usepackage{booktabs}
\usepackage{multirow}
\usepackage{soul}
\usepackage{tabularx}
\usepackage{placeins}
\usepackage{caption}

\usepackage{pifont}
\newcommand{\xmark}{\ding{55}}%
\newcommand{\cmark}{\ding{51}}%

\usepackage{color}

\begin{document}

\title{An Efficient Semi-Automated Scheme for Infrastructure LiDAR Annotation}

\author{Aotian Wu, Pan He$^\dagger$, Xiao Li, Ke Chen, Sanjay Ranka, Anand Rangarajan \\ 
 Modern Artificial Intelligence and Learning Technologies Lab\\ 
Department of Computer and Information Science and Engineering \\
The University of Florida, Gainesville, Florida, USA\\

\thanks{$^\dagger$ indicates corresponding author: Pan He (mybestsonny@gmail.com)}}

\markboth{IEEE Transactions on Intelligent Transportation Systems}%
{Shell \MakeLowercase{\textit{et al.}}: A Sample Article Using IEEEtran.cls for IEEE Journals}


\maketitle

\begin{abstract}

Most existing perception systems rely on sensory data acquired from cameras, which perform poorly in low light and adverse weather conditions. To resolve this limitation, we have witnessed advanced LiDAR sensors become popular in perception tasks in autonomous driving applications. Nevertheless, their usage in traffic monitoring systems is less ubiquitous. We identify two significant obstacles in cost-effectively and efficiently developing such a LiDAR-based traffic monitoring system: (i) public LiDAR datasets are insufficient for supporting perception tasks in infrastructure systems, and (ii) 3D annotations on LiDAR point clouds are time-consuming and expensive. To fill this gap, we present an efficient semi-automated annotation tool that automatically annotates LiDAR sequences with tracking algorithms while offering a fully annotated infrastructure LiDAR dataset---FLORIDA (Florida LiDAR-based Object Recognition and Intelligent Data Annotation)---which will be made publicly available. Our advanced annotation tool seamlessly integrates multi-object tracking (MOT), single-object tracking (SOT), and suitable trajectory post-processing techniques. Specifically, we introduce a human-in-the-loop schema in which annotators recursively fix and refine annotations imperfectly predicted by our tool and incrementally add them to the training dataset to obtain better SOT and MOT models. By repeating the process, we significantly increase the overall annotation speed by $3- 4$ times and obtain better qualitative annotations  than a state-of-the-art annotation tool. The human annotation experiments verify the effectiveness of our annotation tool. In addition, we provide detailed statistics and object detection evaluation results for our dataset in serving as a benchmark for perception tasks at traffic intersections.

\end{abstract}


\section{Introduction}\label{sec:introduction}

Currently, 55 percent of the global population lives in urban areas or cities, which is estimated to increase to 68 percent by 2050. As the world continues to urbanize, we have seen increased investment in building smart traffic infrastructure to achieve the goals of Vision Zero---zero deaths and no serious injuries on roads and streets. For example, the Infrastructure Investment and Jobs Act passed in 2021 by the U.S. government established the new Safe Streets and Roads for All (SS4A) program with an annual budget of  one billion dollars from 2022 to 2026.

\begin{figure}[hbt!]
    \centerline {\includegraphics[width=0.48\textwidth]{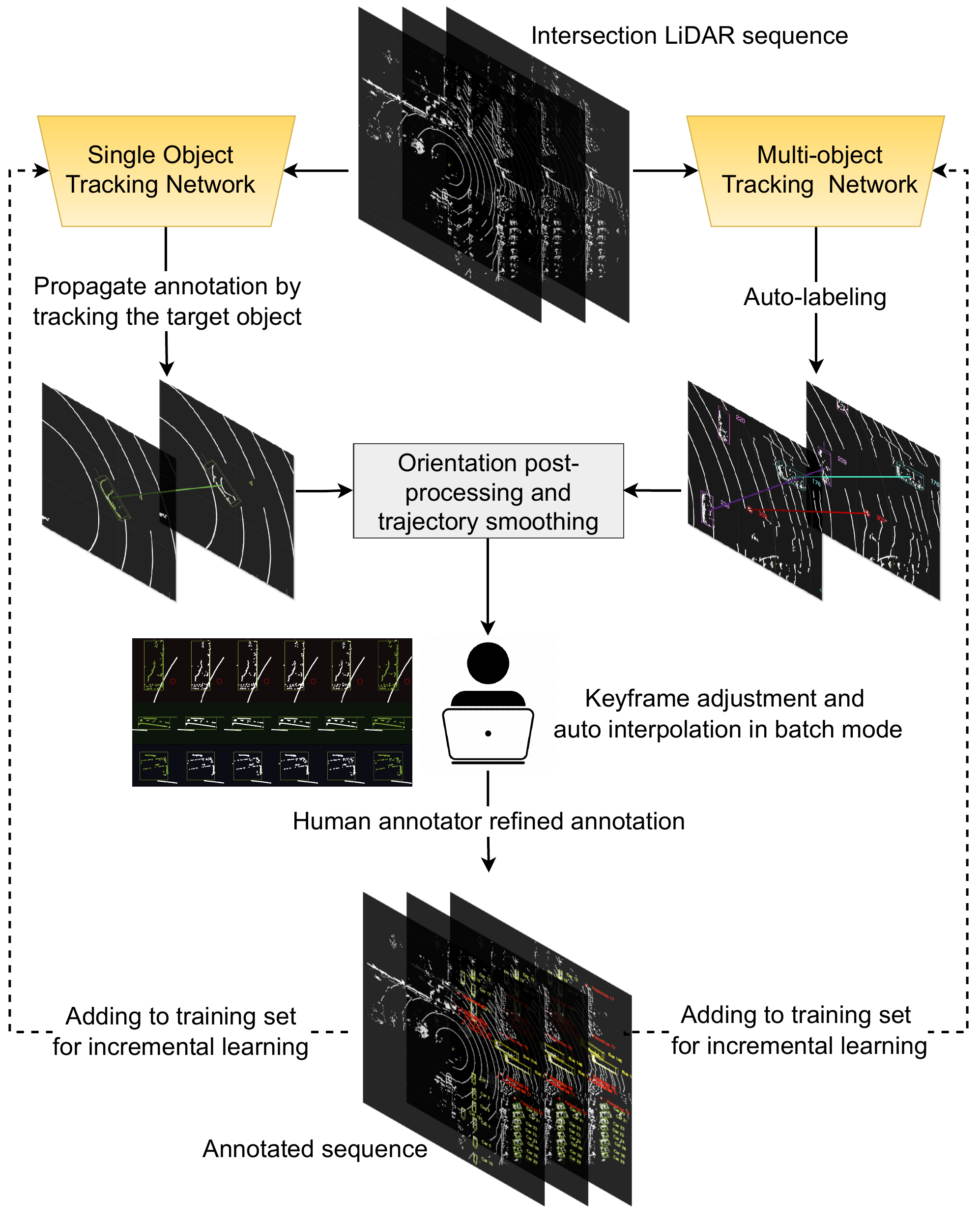}}
    \caption{Overview of the semi-automated annotation pipeline.}
    \label{fig:first_figure}
\end{figure}
Solutions aimed at  Vision Zero goals can be broadly divided into two categories: (i) onboard solutions [e.g., advanced driver assistance systems (ADAS) and autonomous vehicles] that rely on onboard sensing units on vehicles and drones. etc. and (ii) infrastructure solutions (e.g., traffic monitoring systems, traffic lights, speed bumps, streetlamps) that deploy a variety of sensors in transportation infrastructure. Most existing perception systems begin with sensory data acquired from cameras as they provide excellent image/video data streams at an affordable price. However, these solutions suffer from performance drops in low illumination or adverse weather conditions. Moreover, the monocular camera lacks depth information forcing object detection to be confined to 2D. Stereo cameras can obtain depth information via view interpolation but fail to give accurate depth at a distance. Considering the above limitations of cameras, LiDAR---a 3D sensing technology---has received increased attention, especially in creating next-generation infrastructure. By capturing millions of points with precise 3D distance measurements per second through emitting and receiving light pulses (in wavelengths roughly ranging from 900 to 1500nm), LiDAR can support long-range object detection and, in principle, can perform well under various lighting and weather conditions. Due to these characteristics, LiDAR has been widely used in onboard solutions in autonomous driving applications. However, LiDAR-based infrastructure solutions for traffic monitoring systems are still in their infancy.

We identify several significant obstacles while exploring LiDAR-based infrastructure solutions at traffic intersections. To begin with, publicly available LiDAR datasets are, in the main, insufficient for perception tasks in infrastructure systems. Most existing perception tasks in the LiDAR space have relied on public datasets collected from autonomous vehicles in their quest to develop deep learning models for onboard solutions. Despite significant progress \cite{ yurtsever2020survey}, these approaches fail to analyze complex, crowded, and safety-critical scenarios, such as at a busy intersection, due to a limited field of view and heavy occlusion. For these and related reasons, existing onboard solutions are inadequate for supporting the detection of pedestrians, who are more likely to get injured in a traffic accident: (i) popular autonomous driving datasets such as Waymo \cite{sun2020scalability}, NuScenes \cite{caesar2020nuscenes}, and KITTI \cite{geiger2013vision} only provide a limited set of pedestrians for training and evaluation of pedestrian perception algorithms; (ii) pedestrians are small and non-rigid with various poses, making it difficult for sensors to capture; (iii) pedestrians tend to walk in groups, adjust their speed and direction more frequently and unexpectedly (for a safe interpersonal distance), which leads to complex pedestrian behavior and often causing heavy sensor occlusion. On the other hand, infrastructure solutions have an overhead view of traffic and pedestrians with less occlusion. Perception systems in this space offer the promise of a better understanding of challenging and crowded traffic scenarios, leading to more reliability in spotting safety threats. 

A serious challenge for infrastructure LiDAR is that 3D annotations of LiDAR point clouds are time-consuming and expensive. In the course of our initial annotations of an intersection LiDAR dataset, we discovered that annotating and adjusting a single 3D Bounding Box (BBox) around an object is challenging due to its seven degrees of freedom (DoF), namely, the 3D location, 3D size, and heading orientation. Although some annotation tools~\cite{wang2019latte, li2020sustech} are equipped with one-click auto-fitting functions, they fail to accurately annotate under many circumstances, such as when the object is partially occluded, or when the point cloud is sparse. As a result, existing tools require significant effort in data annotation. For example, as stated in a recent pedestrian dataset STCrowd~\cite{cong2022stcrowd}, it took 960 person-hours effort of 20 professional annotators to annotate 219K bounding boxes in the point clouds.

To fill this gap, we present an efficient semi-automated annotation tool that automatically annotates LiDAR sequences with human-in-the-loop initialization and correction. In this work, we construct a fully annotated infrastructure LiDAR dataset that will be made publicly available. Our development is motivated by several key observations. After annotating an object, a common annotation strategy is to propagate the bounding box of the target object to subsequent frames, thereby eliminating the need to label each frame. The strategy is particularly advantageous for 3D data collected at traffic intersections because the size of an object remains constant, e.g., a parked car, or only varies slightly, e.g., a walking pedestrian. Current annotation tools either track objects using Kalman filter-based algorithms~\cite{wang2019latte} or regress the target's movement between two consecutive frames using registration algorithms~\cite{li2020sustech}. The Kalman filter-based approach fails to locate the object precisely and necessitates multiple manual adjustment operations, thereby increasing annotation time. Additionally, the registration algorithm is susceptible to temporary occlusions and tends to lose track of an object after a few frames. Therefore, we seek to use Single Object Tracking (SOT)---a deep learning-based object tracking algorithm---for annotation propagation. Given an object's first-frame annotation, our algorithm can track it robustly in the subsequent frames while maintaining the flexibility of being trained on autonomous driving LiDAR datasets or infrastructure LiDAR datasets. Through extensive experiments, we find that it works well in practice. Furthermore, inspired by the work~\cite{qi2021offboard}, we incorporated a Multi-Object Tracking (MOT) algorithm into our annotation tool. Unlike the SOT, which focuses on independently annotating and refining each object instance via labeling the first frame of each object and propagating it to subsequent frames followed by refinements, the MOT algorithm can automatically detect and track all object instances of a scene in a single shot.  Once it generates the predicted annotation, human annotators may visually inspect and adjust the results. In practice, initial annotations are provided by a trained MOT model. If MOT fails to detect objects, one can annotate its first appearance and utilize an SOT model to propagate. Both SOT and MOT models may not initially give desirable predictions for annotation. Our human-in-the-loop schema allows us to fix and refine imperfectly predicted annotations and improve upon them to recursively obtain better annotations. We show through experiments that the model prediction accuracy is consistently enhanced by adding more qualitative annotations to the training set. As a result, our tool significantly accelerates the overall annotation speed. To summarize, we make the following  contributions:

\begin{itemize}
    \item We develop a semi-automated annotation tool that applies SOT and MOT models while using a human-in-the-loop concept. 
    \item We obtain a large-scale fully-annotated infrastructure LiDAR dataset containing a variety of traffic participants and interesting scenarios. 
    \item We provide baselines for 3D object detection, where the 3D AP for vehicles and pedestrians are 90.66\% and 87.44\%, respectively.
    \item Human annotation experiments demonstrate that our proposed annotation scheme and tool increase the annotation speed of pedestrians and vehicles by approximately a factor of three.
    \item We demonstrate the practical value of this approach and suggest how downstream applications can take advantage of the infrastructure dataset.
\end{itemize}

\section{Background}\label{sec:related_work}
\subsection{3D Single Object Tracking on Point Clouds}
3D single object tracking on point clouds is a relatively new research area. In 2019, SC3D~\cite{giancola2019leveraging} introduced the 3D SOT problem and implemented a Siamese tracker that encodes the target and candidates into embeddings, followed by the cosine similarity measure to determine the best-matching candidate. In addition, it regularized the target embedding by imposing a shape completion loss. P2B~\cite{qi2020p2b} argued that SC3D's candidate generation is either time-consuming or performance-degraded. It then proposed an end-to-end Siamese tracker. Target and search areas are fed to a Pointnet backbone to obtain seeds with features. Then, each seed is projected to a potential target center using Deep Hough voting \cite{qi2019deep}. Finally, P2B clusters the projected target centers and generates the final proposals by choosing those with the highest targetness scores. Multiple successive works ~\cite{zheng2021box, wang2021mlvsnet,hui20213d, zhou2022pttr} are built on top of P2B with additional innovations w.r.t. feature extraction, template and search area feature fusion, and detector heads. BAT~\cite{zheng2021box} proposed a BoxCloud representation that captures the point-to-box relation between object points and their BBoxes. In addition, BAT developed a box-aware feature fusion module to aggregate the features of target points into search area points. MLVSNet~\cite{wang2021mlvsnet} finds that the Hough voting in P2B generates very few vote centers for sparse objects and then proposes multi-level Hough voting as a remedy and a target-guided attention module for feature fusion. In V2B~\cite{hui20213d}, the authors proposed a new voxel-to-BEV detection head. It regresses the target's 3D location in BEV feature maps. PTTR \cite{zhou2022pttr} tracks objects in a coarse-to-fine manner with the help of transformers. It utilized self-attention for template and search area features, respectively, followed by cross-attention for feature fusion, and a generation of coarse prediction builds upon those features. Another lightweight Prediction Refinement Module generates the final predictions. 
\begin{table*}[t!]
\centering
\caption{Comparison of FLORIDA with other popular infrastructure lidar benchmarks.}
\begin{tabular}{@{}llllll@{}}
\toprule
Dataset & \begin{tabular}[c]{@{}l@{}}With crowded \\ pedestrians\end{tabular} & \begin{tabular}[c]{@{}l@{}}Include all traffic \\ participants\end{tabular} & \begin{tabular}[c]{@{}l@{}}Release full dataset \\ with labels\end{tabular} & \begin{tabular}[c]{@{}l@{}}Object detection \\ evaluation\end{tabular} & Annotation method \\ \midrule
Ko-PER & \xmark & \cmark & \cmark & \xmark & Not mentioned \\ \midrule
PedX & \cmark & \xmark & \cmark & \xmark & \begin{tabular}[c]{@{}l@{}}3D model fitting for auto 3D labeling from 2D \\ segmentation and joint location labels\end{tabular} \\ \midrule
IPS 300+ & \cmark & \cmark & \xmark & \cmark & By Datatang Co. Ltd. \\ \midrule
LUMPI & \cmark & \cmark & \xmark & \xmark & \begin{tabular}[c]{@{}l@{}}1) Foreground segmentation using DBSCAN algorithm \\ 2) Annotation propagation using Kalman Filter \\ 3) 3D pose correction using ICP \\ 4) Costly human refinement\end{tabular} \\ \midrule

FLORIDA (Ours) & \cmark & \cmark & \cmark & \cmark & \begin{tabular}[c]{@{}l@{}}1) Auto labeling using MOT \\ 2) Missing object annotation using SOT \\ 3) Pedestrian Orientation auto-correction from \\ moving direction \\ 4) Human refinement in batch mode\end{tabular} \\ \bottomrule
\end{tabular}
\label{tab:comparison}
\end{table*}
The trackers mentioned above all follow the Siamese paradigm and are essentially doing appearance matching between the target and search area. Recently, $M^2$-track~\cite{zheng2022beyond} proposed a new paradigm, namely the motion-centric paradigm. First, it predicts the relative target motion between two consecutive frames. Then it refines the prediction by aggregating the two point clouds with motion compensation to create a denser point cloud. $M^2$-track achieved state-of-the-art performance on multiple benchmarks. In this paper, we adopt $M^2$-track as the SOT model in our annotation tool.

\subsection{3D Multi-Object Tracking on Point Clouds}
The research community initially analyzed the MOT problem in 2D representations, where we track objects in a sequence of images. For 2D MOT, the same objects across frames are associated by appearance and motion cues. For 3D MOT in point clouds, appearance cues become less discriminative because of the sparsity of point clouds and lack of texture information. In contrast, motion cues become more reliable because the scale of an object remains constant, and there are no abrupt movements. Given these characteristics, most of the 3D MOT work employs the tracking-by-detection paradigm and focus on motion modeling for data association. Due to the rapid development of autonomous driving, a variety of LiDAR-based object detectors have been developed and made available, including representative works such as SECOND~\cite{yan2018second}, PointPillars~\cite{lang2019pointpillars}, PointRCNN~\cite{shi2019pointrcnn}, Part$A^2$ Net ~\cite{shi2020points}, CenterNet3D~\cite{yin2021center}, and PVRCNN~\cite{shi2020pv}. For tracking, AB3DMOT~\cite{weng20203d} proposed a baseline approach that adopts the 3D Kalman Filter as the motion model and uses the Hungarian algorithm as the matching strategy. Follow-up work \cite{wu20213d, chiu2021probabilistic} mainly improves upon its data association method and life cycle management strategy. SimpleTrack~\cite{pang2021simpletrack} encapsulates multiple 3D MOT methods (following the  tracking-by-detection paradigm) into a unified framework with four configurable modules, namely detection result pre-processing, data association, motion modeling, and life cycle management. Given its flexibility and simplicity, we employ SimpleTrack as part of our annotation pipeline.

\subsection{Smart Annotation Tools for Point Clouds}
3D BAT~\cite{zimmer20193d} is one of the earliest open-sourced point cloud annotation tools---a web-based application with multi-model data. The annotations for point clouds are automatically projected to different camera views. It also supports interpolation between keyframes to accelerate sequence annotation. LATTE~\cite{wang2019latte} further implemented sensor fusion, smart one-click annotation, and integrated tracking into sequence annotation. LATTE used a clustering algorithm to achieve the one-click annotation to find all points for the target object, estimate the 2D bounding box (BBox), and convert to 3D BBox coordinates based on camera-LIDAR calibration. In addition, LATTE  utilized the Kalman Filter algorithm for tracking objects. SAnE~\cite{arief2020sane} improves one-click annotation by employing a denoising pointwise segmentation strategy that assigns a noise penalty for all boundary locations to better separate nearby objects. SAnE also proposed an improved tracking algorithm, namely a guided tracking algorithm. It consists of 3 stages: greedy search, backtracking, and refinement.

SUSTechPOINTS~\cite{li2020sustech} is one of the best open-source point cloud annotation tools to the best of our knowledge. It has a handy interface for adjusting BBox in single frame or batch mode. Moreover, it implements a collection of functions, such as one-click annotation and annotation propagation. It employs a heuristic registration algorithm that calculates the relative geometric transformation between the target in consecutive frames to propagate the current BBox to subsequent frames. Unfortunately, the registration performance is imperfect, requiring a certain amount of effort in label refinement and correction.  We built our work upon SUSTech POINTS with an improved annotation propagation algorithm. In addition, we extend its functions to include auto-labeling using an MOT tracker, orientation adjustment, trajectory smoothing, etc.

\subsection{Point Cloud Benchmark Datasets}
Many point cloud benchmark datasets focus on autonomous driving applications in which the LiDAR is mounted on moving vehicles. For example, KITTI---a point cloud dataset released in 2013 and now a pioneering vision benchmark \cite{geiger2013vision}---is widely used for evaluating perception tasks. Later, more autonomous driving datasets appeared, comprising more diverse scenes, larger sizes, and more fine-grained annotations. Argoverse \cite{chang2019argoverse}, Nuscenes \cite{caesar2020nuscenes}, and the Waymo Open Dataset\cite{sun2020scalability} are some of the most well-known datasets.

Infrastructure-side point cloud benchmarks, on the other hand, are scarce. To our knowledge, the first infrastructure LiDAR dataset was released in 2014 and is referred to as the Ko-PER Intersection dataset \cite {strigel2014ko}. It deploys 14 SICK LD-MRS 8-layer research laser scanners to a four-way intersection in Aschaffenburg, Germany. Later in 2021, IPS300+ \cite{wang2021ips300+} released a high-density intersection dataset. It installs two 80-beam Robosense Ruby-Lite LiDAR scanners at the diagonal of a 4-way intersection. The two LIDAR cameras are calibrated and cover the entire intersection. However, only a total of 600-frame annotations are made available. Recently,  LUMPI \cite{busch2022lumpi} proposed a multi-perspective intersection dataset in Hanover, Germany. It deployed three cameras and 5 LiDARs to cover the intersection with dense point clouds. And a total of 90K point clouds have been released. However, their labels are unavailable as of November 21, 2022.


Our proposed dataset is collected at a busy intersection near the University of Florida campus, comprising crowded vehicles, pedestrians, and a parking lot. We captured sequences covering diverse traffic behaviors such as pedestrian jaywalking, near-misses, vehicles lining up on the crosswalk, causing pedestrians to take detours, and people exiting vehicles while waiting at a red light. We demonstrated through our FLORIDA dataset that a single LiDAR can sufficiently capture most of the intersection traffic, except for a 5-meter blind spot beneath the LiDAR. And our semi-automated annotation algorithm performs well under the LiDAR-only setting.

\section{Methodology}\label{sec:method}
\algnewcommand\Input{\item[\textbf{Input:}]}%
\algnewcommand\Output{\item[\textbf{Output:}]}%

This section introduces the collected dataset, detailed statistics regarding performance and a qualitative comparison with other infrastructure LiDAR datasets. Then, we present an overview of the proposed semi-automated annotation scheme, followed by an explanation of the utilized deep-learning-based models. Finally, we discuss the pre- and post-processing algorithms designed to further improve annotation speed.


\subsection{The FLORIDA Dataset}
\label{subsection:uni13}

\subsubsection{Data collection}
We collected the dataset at a busy intersection---West University Avenue \& Northwest 17th Street, Gainesville, FL---near the campus of the University of Florida. The LiDAR camera is mounted on a 5-meter post at the intersection. We used a Velodyne VLP-32C LiDAR with 32 channels, a 200-meter range, $+15$° to $-25$° vertical field of view (FOV), and 360° horizontal FOV. We manually selected 11 sequences, some of which included crowded pedestrians, abnormal behaviors, or near-misses. Henceforth, we refer to our dataset as \textbf{FLORIDA}---\textbf{F}lorida \textbf{L}iDAR-based \textbf{O}bject \textbf{R}ecognition and \textbf{I}ntelligent \textbf{D}ata \textbf{A}nnotation.

\subsubsection{Dataset statistics and characteristics}
\begin{figure*}
    \centering
    \includegraphics[width=0.9\textwidth]{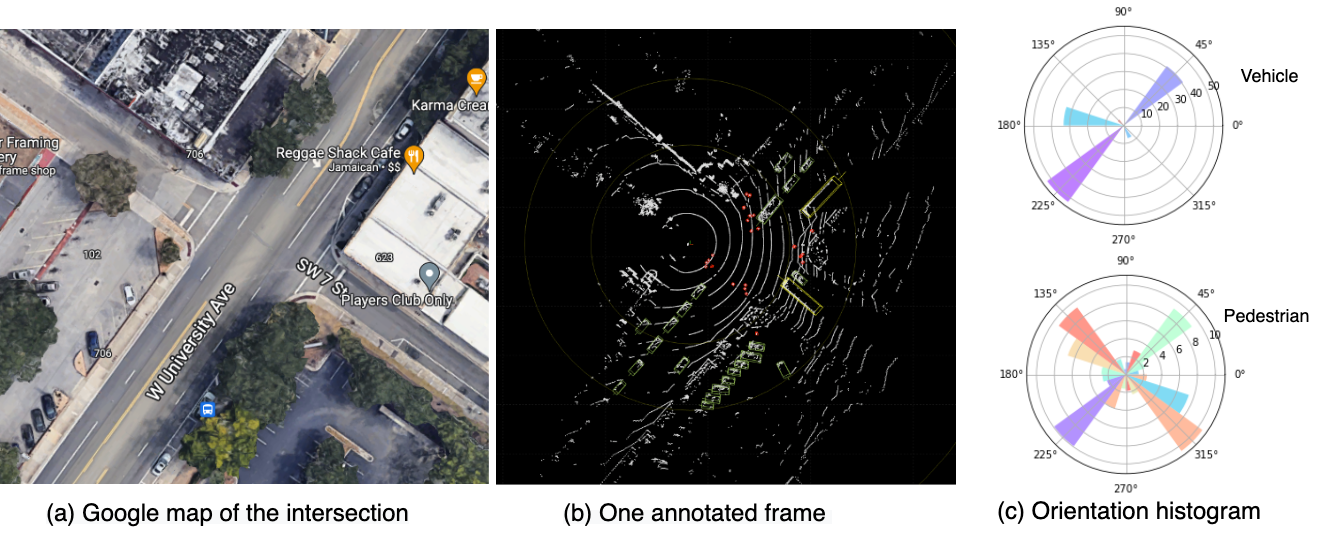}
    \caption{We collect crowded pedestrian sequences from a LiDAR installed at a busy intersection  --- West University Avenue \& Northwest 17th Street, Gainesville, FL, near the campus of the University of Florida. (c) is the orientation histogram of vehicles and pedestrians, and the count numbers in the images are measured in thousands.}
    \label{fig:demonstration}
\end{figure*}

\begin{figure*}
    \centering
    \includegraphics[width=0.88\textwidth]{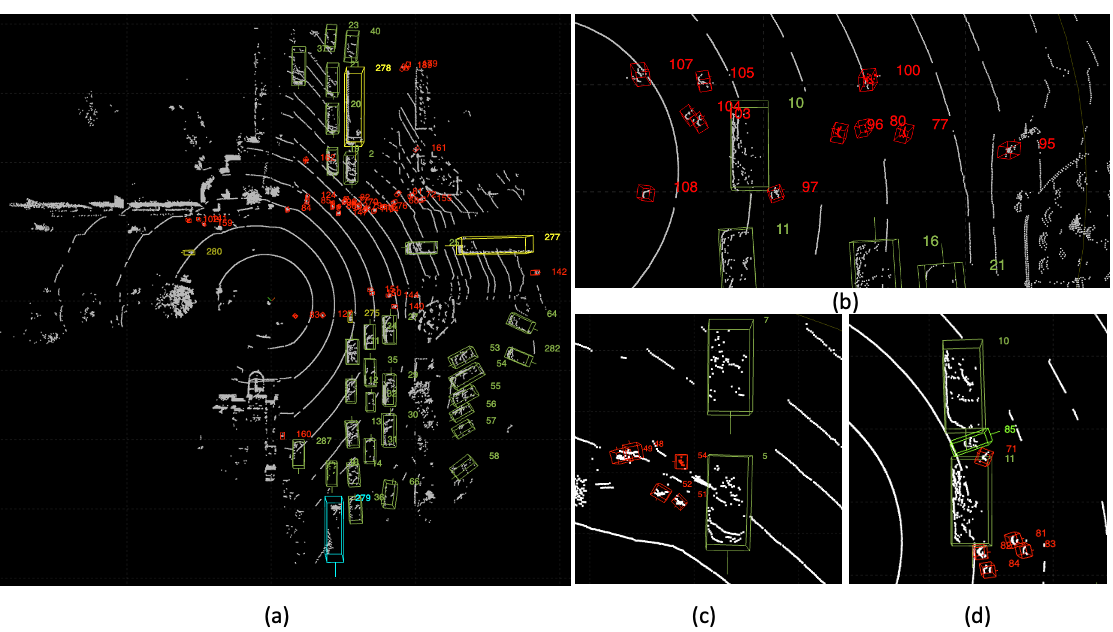}
    \caption{(a) A sample of a crowded scene. (b) A vehicle stopped at a pedestrian crosswalk. (c) People exiting a vehicle stopped at a red light. (d) A cyclist and a pedestrian passing through a small gap between two cars.}
    \label{fig:samples}
\end{figure*}

 As shown in Table~\ref{tab:statistics} and Figure~\ref{fig:demonstration}~(c), we first summarize the statistics for all categories and display the orientation histogram of vehicles and pedestrians, respectively. The orientation histogram indicates that most vehicles move in a 45°/225° direction, corresponding to West University Avenue. The 165° direction comes from a parking lot where most vehicles park in parallel. For pedestrians, most of them cross the streets in the zebra-crossings, resulting in spikes in 45°, 225°, 135°, and 325° directions. As it is difficult to determine the orientations of pedestrians from the point cloud when they are waiting to cross the intersection, we utilized a heuristic approach, which we detail in Section~\ref{section:post_processing}. As depicted in Figure~\ref{fig:samples}, the FLORIDA dataset captures crowded pedestrians and vehicles and several abnormal behaviors, which is beneficial for training and evaluating object detectors and trackers under challenging conditions, such as scenes with crowds with numerous occlusions.  

The comparisons with some of the popular infrastructure LiDAR datasets are summarized in Table~\ref{tab:comparison}. In brief, previous datasets either did not release the full dataset labels or did not report the evaluation performance, such as object detection. To the best of our knowledge, FLORIDA is the first dataset to include crowded pedestrians and diverse traffic participants which will be fully released and can be openly evaluated for object detection. In addition, we compared the annotation approaches of all datasets. LUMPI is most comparable because they only annotate on LiDAR point clouds. Compared to LUMPI, our annotation approach employs trained deep-learning models that improve the accuracy and robustness of auto-labeling and provide more assistance to human annotators for post-correction and refinement.

 \begin{table*}
\caption{The statistics of FLORIDA}
\centering
\begin{tabular}[width=\textwidth]{@{}lrrrrrrr@{}}
\toprule
Class & Vehicle & Pedestrian & Cyclist & Motorcycle & Bus & Truck \\ \midrule
The total number of instances & 143,941 & 80,220 & 999 & 17,397 & 4,170 & 2,640   \\
The average number of instances per frame & 21.81 & 12.15 & 0.15 & 2.64 & 0.63 & 0.40  \\
The maximal number of instances per frame & 38 & 34 & 2 & 7 & 3 & 3  \\ \bottomrule
\end{tabular}
\label{tab:statistics}
\end{table*}



\subsection{Overview of the Semi-automated Annotation Algorithm }

A common strategy for annotating a new dataset is to annotate object instances one by one, from their first appearance to their exit from the scene. Typically, an annotation tool can leverage a tracking algorithm to track and annotate an object across multiple frames. In a similar vein, we propose to exploit a state-of-the-art deep-learning SOT tracker \cite{zheng2022beyond} to propagate annotations: we begin by providing the initial annotation of an object with a proper one-click function and then propagating the annotation across subsequent frames (e.g., up to 100 frames) using the SOT tracker. Of course, the auto-generated BBoxes might be imperfect; therefore, manual annotator refinement is necessary. Following \cite{li2020sustech}, we leverage the function of batch-mode editing in which adjusting keyframes' annotations could trigger the interpolation of intermediate frames, which is proven to reduce refinement effort. 

We further automate the annotation using a trained MOT, which generates tracklets for all objects.
In contrast to SOT, MOT does not require first-frame annotation for each object. The MOT is iteratively trained. In the beginning, it is trained on one fully annotated sequence. As more sequences are annotated, its training set is expanded such that the detection and tracking accuracy improves accordingly. Nevertheless, the MOT algorithm will still miss some objects or provide imprecise annotations.  The annotator will then check each tracklet and make necessary adjustments. Additionally, the SOT can be utilized as a remedy for objects missed by MOT. 



The aforementioned annotation scheme is not limited to static LiDAR settings; it can also be used to speed up the annotation for onboard LiDAR datasets. Specific to the static LiDAR setting, we developed a collection of pre- and post-processing algorithms: ground height estimation, trajectory smoothing, orientation post-processing, and static object BBox averaging. We now describe these methods in detail. 

\subsection{Annotation Propagation by Single Object Tracking}

Given a point cloud sequence and a BBox of an object in the first frame as the input of a 3D SOT tracker, we aim to locate the same object in a sequence of frames. Specifically, given a point cloud sequence $\{P^t\}_{t=1}^{T}$ of frame length $T$ and the 3D BBox $B^1 \in \mathbb{R}^7$ of one object, parameterized by its location in 3D coordinates, height, length, width, and heading direction, at the first frame, a SOT tracker aims to find all 3D BBoxes of the object in subsequent frames denoted as $\{B^t\}_{t=2}^{T}$.

In our setting, the annotator provides the initial BBox annotation, followed by the trained SOT tracker locating the object frame-by-frame. Because objects move continuously in 3D, their locations in two consecutive frames are close; therefore, the search area can be $K$ meters around the object's last location. $K$ is a hyper-parameter determined by the object's velocity and the frame rate of the LiDAR data. Following \cite{zheng2022beyond} and taking our static LiDAR setting into account, we empirically set $K$ to $2$ for vehicles and $0.5$ for pedestrians.

To regress the position offset of the object between two consecutive frames, we resort to the state-of-the-art 3D SOT model---$M^2$-Track \cite{zheng2022beyond}. It proposes a two-stage motion-centric paradigm in which the motion transformation between the same objects in two frames is first regressed, followed by a refinement based on the merged point cloud in two frames. In detail, $M^2$-Track initially segments the target points in two frames using a trained semantic segmentation network. Then a motion vector $M = (\delta x, \delta y, \delta z, \delta \theta)$ is regressed by a motion estimation network, where $\delta x, \delta y, \delta z$ represent the location offsets, and $\delta \theta$ represents the heading direction angle offset. Adding the motion vector $M$ to $B^{t-1}$ gives us a coarsely predicted BBox $\hat{B}^{t}$. In the second stage, $M^2$-Track refines $\hat{B}^{t}$  by regressing a small relative offset and producing the final prediction $B^t$. Specifically, $M^2$-Track aggregates the previous frame point clouds $P^{t-1}$ into the current point cloud $P^{t}$, compensating motion using the predicted $M$, resulting in a denser point cloud $\tilde{P^{t}}$. Another regression network is applied to $\tilde{P^{t}}$ to produce the refined BBox $B^t$.
 We integrate the SOT model into SUSTechPOINTS \cite{li2020sustech}---a popular open-source annotation tool for point clouds---by replacing the original auto-labeling function of SUSTechPOINTS with $M^2$-Track, resulting in a more robust function for handling  occlusion and sparsity and producing better accuracy for deformable objects like pedestrians. We utilize the SOT model to select a BBox to propagate to subsequent $N$ frames, which returns $N$ BBox predictions and displays them on the annotation tool. The parameter $N$ can be changed by the annotator, depending on the scenario. For example, more adjustments from the annotator will be necessary when there is heavy occlusion or in a crowded area. Therefore, a smaller value for $N$ is more practical in such a situation. By default, we set it to a fixed number (i.e., $N=100$) for typical cases. The annotator can switch to batch processing mode, where adjusting the keyframes will trigger interpolation for middle frames, which is beneficial in speeding up the annotation. Next, if the object is still visible after $N$ frames, one can adjust the last-frame annotation and continue propagating the annotation to subsequent frames. To harmonize the SOT algorithm and interpolation, we set one annotation out of every ten as a keyframe such that the annotator can quickly refine the annotation by adjusting keyframes alone most of the time. Adjusting keyframes may be insufficient for turning vehicles, as the orientation change is non-linear. In this case, we can refine some annotations that are not keyframes. Once refined, a non-keyframe will change to a keyframe and accordingly trigger the interpolation.

\subsection{Auto-annotation by Multi-Object Tracker}

Given a point cloud sequence, the goal of MOT is to localize and identify all objects in the sequence. Formally, given point cloud sequence $\{P^i\}_{t=1}^T$, the MOT finds the BBoxes of all objects $\{\{B^t_j\}_{t=1}^{T}\}_{j=1}^{N^t}$, where $N^t$ is the number of objects in frame $P^t$. Note that $N^t$ varies over frames, as objects may enter or exit the scene at different times.

In our annotation scheme, an MOT model automatically generates tracklets for all objects in the scene. To achieve this, we follow a tracking-by-detection paradigm, where we detect all objects via a detector frame-by-frame and then use the tracker to associate boxes for the same object across frames. 

We apply CenterPoint~\cite{yin2021center} for multi-object detection. It detects objects as key points and then regresses their other attributes, namely 3D location, 3D size, and 1D heading orientation. CenterPoint consists of a standard 3D backbone, a center heatmap head, and regression heads. The 3D backbone extracts bird-eye-view (BEV) feature maps fed to the heads to generate predictions. The head produces keypoint heatmaps, where each heatmap peak corresponds to a predicted object center. And another regression head regresses other properties for predicted key points, such as BBox sizes and orientations. We followed OpenPCDet's CenterPoint implementation. The readers can find more details about model architecture, training strategy, and model implementation in CenterPoint~\cite{yin2021center}.

Given predicted boxes in each frame, the next stage is associating the BBox of the same object across frames, producing tracklets of objects. To this end, we adopted SimpleTrack~\cite{pang2021simpletrack}, a top-performing multi-object tracking approach. Following the "tracking-by-detection" paradigm, SimpleTrack unifies the 3D MOT methods into a general framework. The framework consists of four main components: (i) detection pre-processing, (ii) BBox association across frames, (iii) object motion modeling, and (iv) tracklet lifecycle management. Given multiple options in each component, we adopted those matching our dataset's characteristics. The pre-processing module mainly processes the raw detection predictions into a cleaner input to the tracker. We follow SimpleTrack to apply a stricter non-maximum suppression (NMS) to the raw detection predictions to preserve recall while improving precision. It effectively removes low-confidence detections that overlap with others while preserving low-confidence detections likely from sparse or occluded regions. For motion modeling, we adopted the Kalman Filter, which predicts the location of an object with increasing precision in the next frame. The Kalman Filter performs exceptionally well on infrastructure datasets because the LiDAR is stationary, resulting in longer tracklets and no abrupt motions. The predicted location from the motion model is then used as a proposal to associate with detections in the next frame. Next, for BBoxes association across frames, we view the problem as a bipartite matching problem and employ the well-known Hungarian algorithm \cite{jonker1987shortest}. As objects enter and exit the LiDAR's field of view at different times, the life cycle of tracklets needs to be carefully maintained. Following SimpleTrack, we adopt the "two-stage association" strategy. The detection score threshold is higher in the first stage than in the second. The first stage ensures tracking precision, while the second stage extends the life of tracklets in occluded or sparse regions, thereby reducing the number of ID switches.

We further post-process the generated tracks with heuristic rules. First, we remove the tracklets that are too short because they are likely to be false positives. Second, we filter the tracklets whose speed is outside a reasonable range. For instance, the typical walking speed of a pedestrian is less than 2 meters per second. Therefore, predicted pedestrian tracklets with a higher average speed are more likely to be cyclists or motorcycles. Finally, because our dataset contains many parked cars, the bounding boxes of such tracklets vary slightly from frame to frame. Therefore, we average them to generate more accurate annotations.

There are cases where the MOT model makes mistakes. For instance, we notice missing detection, incorrect detection, track ID switches, reversed orientation, etc. We further develop functions to assist annotators in quickly correcting errors. We leverage the SOT model for missing detection to propagate annotations for completion. For incorrect detection, annotators could delete all annotations for a given ID. To handle track ID switches, annotators could correct the ID where the switch happens and synchronize the change to the following frames. Lastly, they could correct the reversed orientation via a single one-click or batch correction in batch mode.

\subsection {Pre-processing and Post-processing Algorithms} 
\subsubsection{Trajectory smoothing and orientation post-processing} \label{section:post_processing}

When annotating pedestrians, we find it particularly challenging to determine their orientation from a single frame. For example, the point cloud on a pedestrian could be very sparse and incomplete. Often, the annotator has to examine the sequence surrounding the current frame to determine the orientation of a pedestrian based on movement. Therefore, we developed an orientation post-processing algorithm that imitates the annotator's behavior and significantly reduces the pedestrian annotation time. Specifically, after annotating a sequence, we first smooth the trajectory using a cubic smoothing spline algorithm \cite{pollock1993smoothing} and then set the orientation at each timestamp as the pedestrian's moving direction. It is more tricky to set the orientation for stationary pedestrians. Therefore, we adopt a heuristic strategy: if the pedestrian starts moving later in the sequence, we set the orientation to match the direction of movement; otherwise, the orientation remains the same as the pedestrian's initial orientation. 



\subsubsection{Ground height estimation}
For small objects (i.e., pedestrians and cyclists) that are too close or too far from the LiDAR's center, there are only a few points on each object, and it is ambiguous to determine the object's $z$ value, i.e., height. The ground height information is helpful in such cases, as it allows us to better spot objects on the ground using SOT and MOT algorithms. To obtain the ground height, we manually segment the ground points using Point Cloud Labeler~\cite{behley2019iccv}. Next, the ground points are interpolated into grids using the \texttt{LinearNDInterpolator} from the python SciPy library. However, interpolation does not work well for distant regions with sparse data points. To cover these regions, we estimate a ground plane given all segmented ground points using the RANSAC algorithm \cite{fischler1981random}. The interpolation captures subtle differences in ground height, such as the sidewalk being slightly higher than the road. Additionally, the ground plane captures the intersection's general elevation or the LiDAR's slight tilt angle. Note that the ground height of an intersection only needs to be estimated once.

\section{Experiments}\label{sec:experiment}

In this section, we conduct several experiments to demonstrate the FLORIDA dataset's quality and the annotation scheme's usefulness. We evaluate the speed and accuracy of our developed annotation tool in Section~\ref{subsection:annotator_experiment}. Section~\ref{sec:3d_detection_exp} presents baseline detection results with a study of the trade-off between annotation quantity and detection accuracy. Section~\ref{sec:time_reduction} illustrates the improvement in annotation speed as more data is annotated. Finally, Section~\ref{sec:application} gives an example of a downstream application based on this work.

\subsection{Annotator experiments}\label{subsection:annotator_experiment}
One straightforward way to evaluate the efficiency and accuracy of an annotation tool is to conduct a human annotation experiment. Therefore, we record the annotation time of four annotators and evaluate their annotation quality. As annotators' annotation speed may vary, we conduct the experiment with two trained and two untrained annotators and separately report their average annotation times. 

\begin{figure}
    \centerline {\includegraphics[width=0.45\textwidth]{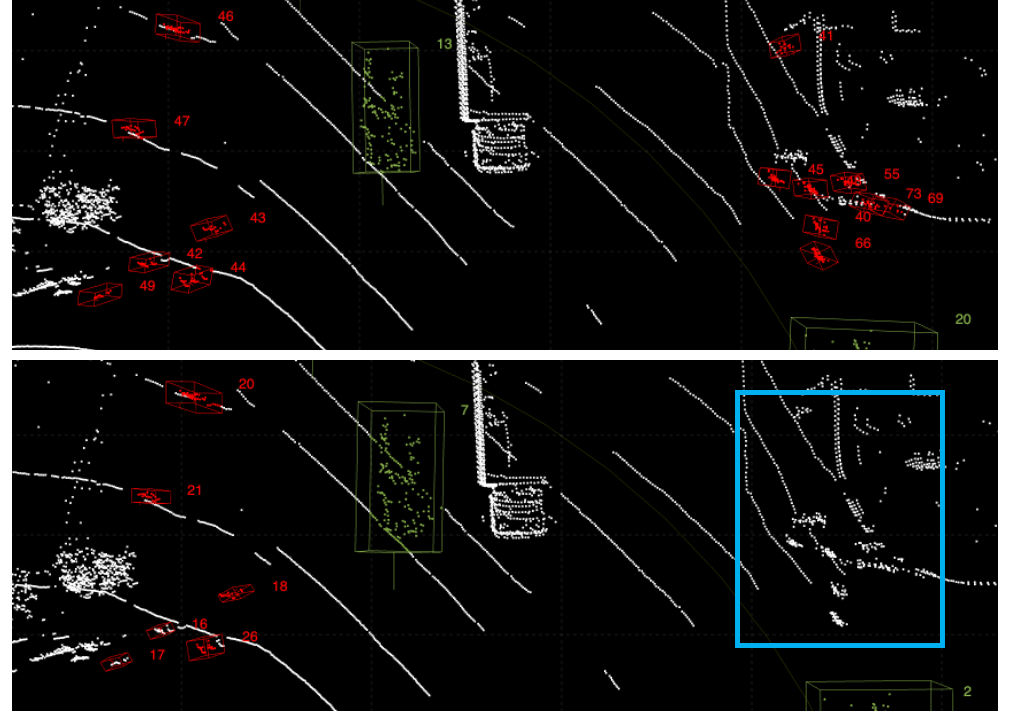}}
    \caption{Example of annotations from an untrained annotator using our tool versus SUSTechPOINTS. The top one is annotated using our tool, while the bottom is annotated using SUSTechPOINTS. The pedestrians within the blue box are not recognized by the annotator.}
    \label{fig:missed_anno}
\end{figure}

We select a 200-frame LiDAR sequence with crowded vehicles and pedestrians and ask annotators to label the same sequence using two different annotation tools---SUSTechPOINTS and ours. Ground truth labels are annotated and double-checked by an experienced annotator. When annotating the ground truth, we verify the annotation by observing a longer sequence. The annotation efficiency is measured by the average annotation time, and the annotation accuracy is measured by the average $F_1$ score. We consider an annotation BBox to be accurate if the Intersection over Union (IoU) with a ground truth BBox exceeds a threshold. As the tightness of BBoxes differs between annotators, in the experiment, we set the IoU threshold at 0.3. Table~\ref{tab:annotation_efficiency} summarizes our tool's annotation efficiency and accuracy against SUSTechPOINTS. It shows that our annotation tool nearly quadruples the speed of annotation for both trained and untrained annotators. Meanwhile, our tool's annotation quality is also better, especially for pedestrians. The main reason is that the MOT algorithm provides a template for the annotator, which largely improves the recall for pedestrians. As shown in Figure~\ref{fig:missed_anno}, the annotator using SUSTechPOINTS does not recognize the pedestrians waiting to cross the street as human annotators recognize objects primarily based on motion, whereas the pedestrians in the blue box are stationary. On the other hand, our MOT algorithm enables the annotator to recognize and accurately annotate these pedestrians.

\begin{figure}
    
    \includegraphics[width=0.45\textwidth]{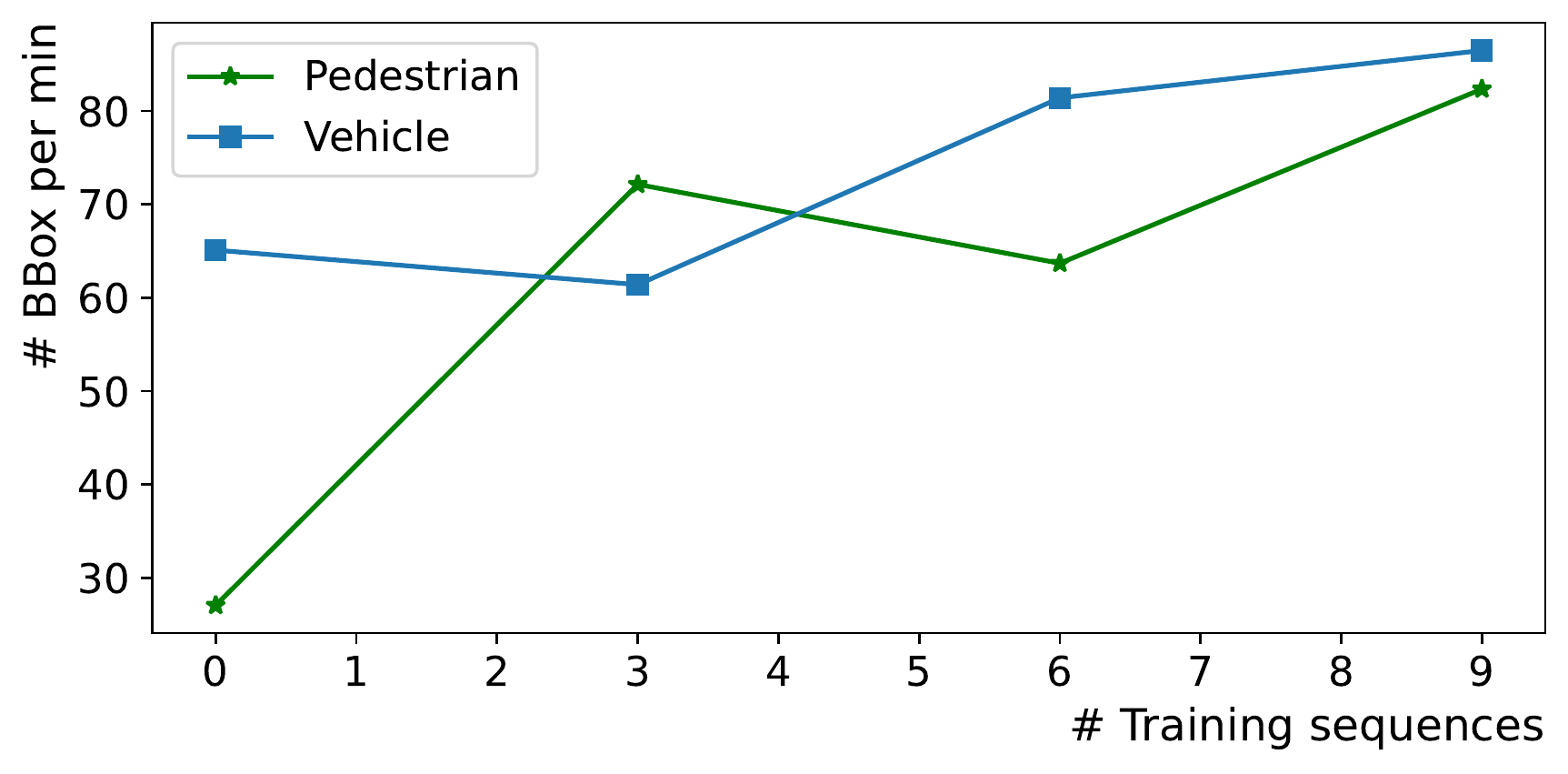}
    \caption{Annotation speed improvement as the MOT is trained on more sequences. \# training sequences = 0 means that we only use SOT for annotation.}
    \label{fig:anno_speed}
\end{figure}

\begin{table}
\centering
\caption{3D object detection result on FLORIDA dataset.}
\begin{tabular}{@{}lllll@{}}
\toprule
 & \begin{tabular}[c]{@{}l@{}}Vehicle\\ IoU = 0.7\end{tabular} & \begin{tabular}[c]{@{}l@{}}Pedestrian\\ IoU = 0.5\end{tabular} & \begin{tabular}[c]{@{}l@{}}Motorcycle\\ IoU = 0.5\end{tabular} & \begin{tabular}[c]{@{}l@{}}Bus\\ IoU = 0.7\end{tabular} \\ \midrule
3D AP (\%) & 90.66 & 87.44 & 82.32 & 91.99 \\
BEV AP (\%) & 96.62 & 87.76 & 96.96 & 95.17 \\ \bottomrule
\end{tabular}
\label{tab:det_ap}
\end{table}

\subsection{3D detection in the FLORIDA Dataset}
\label{sec:3d_detection_exp}
In Table~\ref{tab:det_ap}, we show the 3D detection results for four categories on two 600-frame validation sequences. The validation sequences are collected on different days, without any days overlapping with the training sequences. We use the 3D Average Precision (3D AP) and Bird's Eye View Average Precision (BEV AP) as evaluation metrics, as defined by the KITTI benchmark~\cite{geiger2013vision}. We employ lower  Intersection over Union (IoU) thresholds for smaller objects, such as Pedestrian and Motorcycle, and higher IoU thresholds for larger objects, such as Vehicle and Bus. Truck and Cyclist are annotated, but there are insufficient instances for evaluation, and therefore they are omitted from the table. We investigate the improvement of the detector's AP as more training data is gradually added. As shown in Figure~\ref{fig:ap_growth}, training on 3 600-frame sequences already gives a reasonably good result, whereas the AP improvement from 3 to 6, and  6 to 9 are less significant. Therefore, given the detection accuracy requirement for different downstream tasks, one can vary the amount of annotation.


\begin{table*}
	\begin{minipage}{0.61\textwidth}
            \centering
		\caption{Comparative analysis of annotation speed and precision between SUSTechPOINTS and ours. Our annotation tool speeds up the annotation by more than $3 \times$ while improving the annotation quality, especially for pedestrians.}
		\label{tab:annotation_efficiency}
		\centering
		\begin{tabular}{@{}cccccrr@{}}
            \toprule
            \multirow{2}{*}{} & \multicolumn{2}{c}{SUSTechPOINTS} & \multicolumn{2}{c}{Ours} & \multicolumn{2}{c}{Our Improvement}  \\
             & Car & Pedestrian & Car & Pedestrian  & Car & Pedestrian\\ \midrule
            Avg time-trained (min) & 80.5 & 119.5 & 21.0 & 31.0 &  $\times$3.8 & $\times$3.9 \\
            Avg time-untrained (min) & 152.0 & 185.0 & 34.5  & 47.5 & $\times$4.4 & $\times$3.9 \\
            Avg F-1 score-trained (\%) & 91.0 & 80.6 & 97.8  & 96.4 &  +6.8   & +15.8   \\
            Avg F-1 score-untrained (\%) & 92.5 & 68.0 & 97.6  & 96.8 &  +5.1   & +28.8   \\
            \bottomrule
            \end{tabular}
	\end{minipage}\hfill
	\begin{minipage}{0.37\textwidth}
		\centering
                \includegraphics[width=\textwidth]{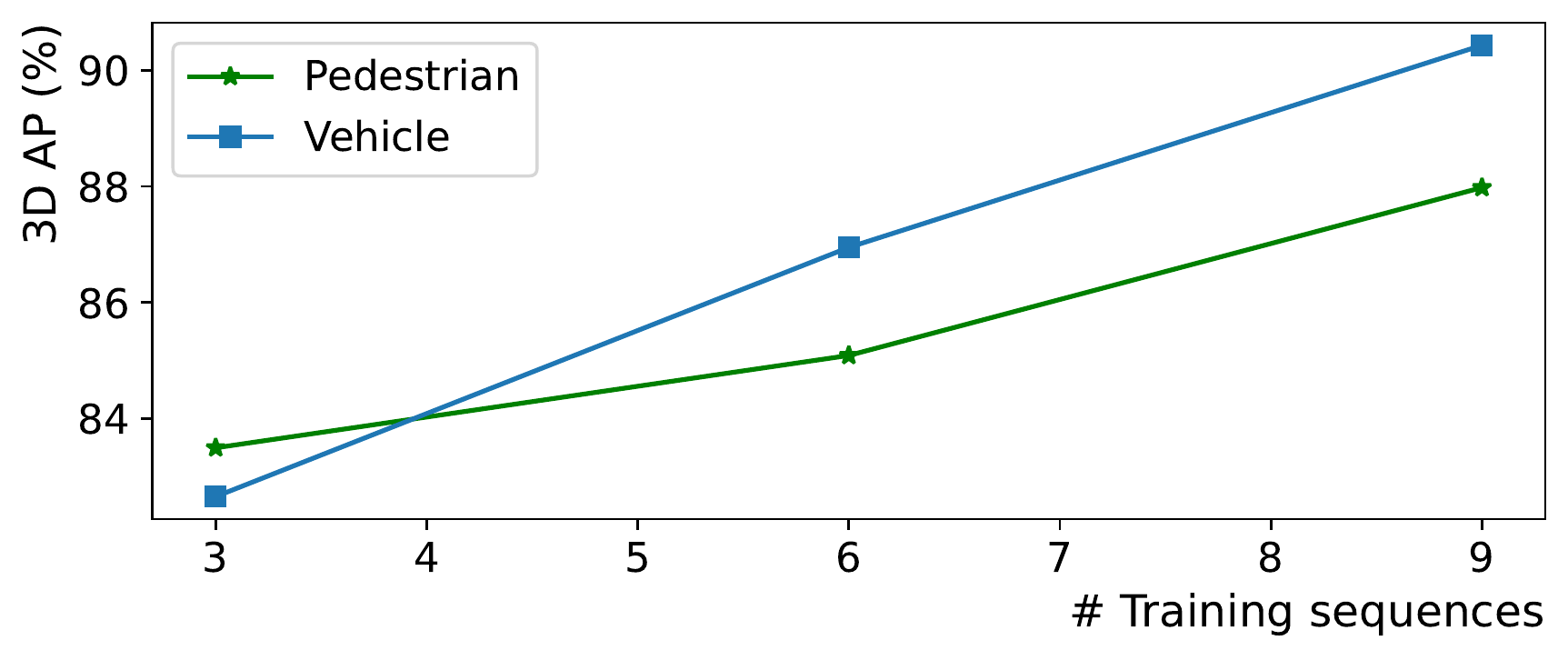}
                \captionof{figure}{3D AP varying the amount of training data. Training on 3 sequences already produces a decent result. The AP further improves when training on more sequences.}
                \label{fig:ap_growth}
	\end{minipage}
\end{table*}

\subsection{Annotation time reduction with training on more data}
\label{sec:time_reduction}
As shown in Figure~\ref{fig:anno_speed}, we record the annotation time for Vehicle and Pedestrian in the FLORIDA dataset to demonstrate the effectiveness of the human-in-the-loop concept. We train a new model for every three 600-frame sequences and calculate the average number of BBoxes per minute to measure annotation speed. As object density and moving patterns vary across different sequences (with the annotation of the crowded scene being more challenging), the resulting data points are fuzzy. However, we still observe a clear trend of increasing annotation speed. For Vehicle, annotation propagation with SOT and batch-mode interpolation already provide high annotation speed. For Pedestrian, MOT significantly increases speed. Through the experiment, the MOT model trained on three sequences increases the annotation speed from 27.05 to 72.13 BBoxes/min.

\subsection{Application to traffic monitoring systems}
\label{sec:application}
We demonstrate the effectiveness of the FLORIDA dataset with our semi-automated annotation suite through a downstream use case. We integrate the predicted object trajectories into a web-based visual analytics system, where one can check all trajectories in a given time period, obtain count statistics for traffic participants, observe abnormal behaviors, etc. Compared with video sensors, LIDAR performs well regardless of the lighting conditions, thereby enhancing the safety of intersections. 

\begin{figure}[t!]
    \centering
    \includegraphics[width=0.45\textwidth]{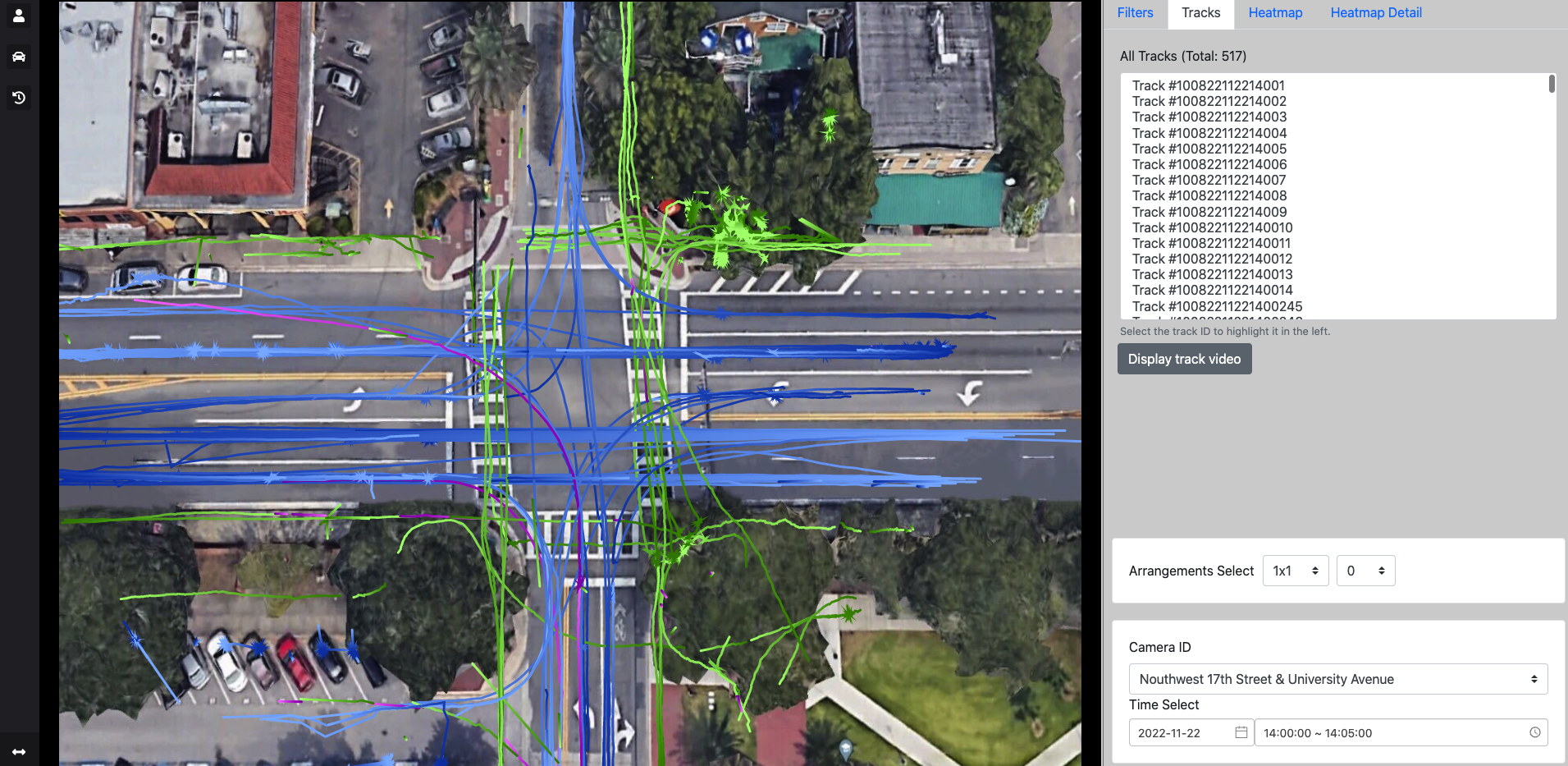}
    \caption{Application of our work on a web-based intersection traffic monitoring system. The trajectories of different types of traffic participants are shown in different colors. One can inspect all the trajectories and analytic statistics of a given time period.}
    \label{fig:application}
\end{figure}

\section{Conclusion and Future Direction}\label{sec:conclusion}


In this paper, we have developed a semi-automated annotation tool that applies SOT and MOT models integrated with the human-in-the-loop schema for speeding up data annotation of challenging intersection LiDAR datasets. We verify its effectiveness via conducting human annotator experiments and reporting qualitative and quantitative results on object detection. Our developed tool supports the creation of achievable and affordable LiDAR-based traffic monitoring systems. Besides, we have introduced a fully-annotated infrastructure LiDAR perception dataset---FLORIDA---consisting of diverse and crowded traffic participants and exciting traffic scenarios, to facilitate research on infrastructure-based object perception. In future work, we aim to enrich the dataset with video cameras and reduce the annotation time given the additional appearance information. We want to study how transfer learning from our dataset can benefit training a model on new scenes, leading to an even faster setup time for a new intersection or road segment.


 \section*{Acknowledgments}
This work is supported by NSF CNS 1922782, by the Florida Dept. of Transportation (FDOT)
and FDOT District 5. The opinions, findings and conclusions expressed in this publication are those
of the author(s) and not necessarily those of the Florida Department of Transportation or the National
Science Foundation. 
\bibliographystyle{IEEEtran}
\bibliography{ref.bib}
\vspace{-1.2cm}
\begin{IEEEbiography}[{\includegraphics[width=0.8in,height=1in,clip]{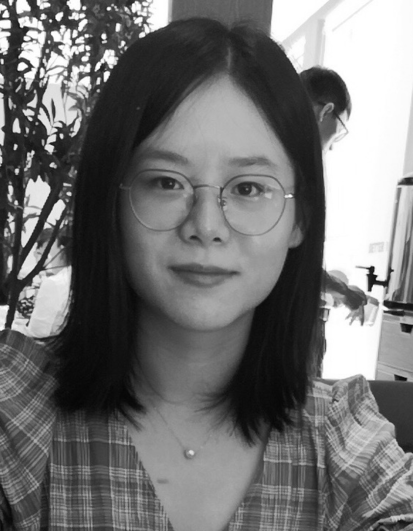}}]%
{Aotian Wu} is currently a Ph.D. student in the Department of Computer \& Information Science \& Engineering, University of Florida , Gainesville, FL. Her research interests include deep learning and computer vision.
\end{IEEEbiography}

\vspace{-1.65cm}
\begin{IEEEbiography}[{\includegraphics[width=0.8in,height=1in,clip]{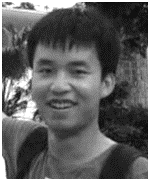}}]%
{Pan He} is currently a Ph.D. student in the Department of Computer \& Information Science \& Engineering, University of Florida , Gainesville, FL. His research interests include deep learning and computer vision.
\end{IEEEbiography}
\vspace{-1.65cm}

\begin{IEEEbiography}[{\includegraphics[width=0.8in,height=1in,clip]{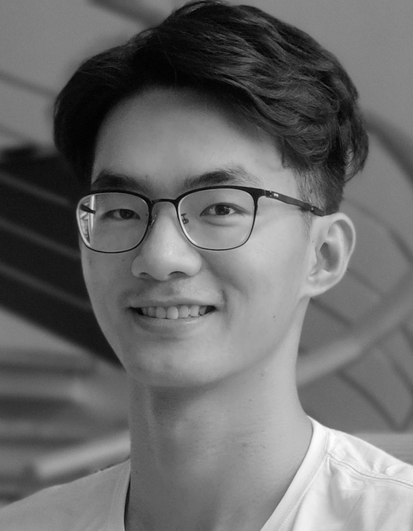}}]%
{Xiao Li} is currently a Ph.D. student in the Department of Computer \& Information Science \& Engineering, University of Florida , Gainesville, FL. His research interests include deep learning and computer vision.
\end{IEEEbiography}
\vspace{-1.65cm}

\begin{IEEEbiography}[{\includegraphics[width=0.8in,height=1in,clip]{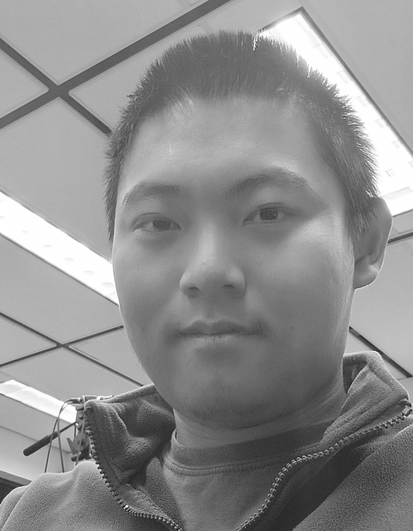}}]%
{Ke Chen} is currently a Ph.D. student in the Department of Computer \& Information Science \& Engineering, University of Florida , Gainesville, FL. His research interests include deep learning and computer vision.
\end{IEEEbiography}
\vspace{-1.65cm}

\begin{IEEEbiography}[{\includegraphics[width=0.8in,height=1in,clip]{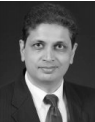}}]%
{Sanjay Ranka} is a Distinguished Professor in the Department of Computer Information Science and Engineering at the University of Florida. His current research is on developing algorithms and software using Machine Learning, Internet of Things, GPU Computing, and Cloud Computing for solving applications in Transportation and Health Care. He is a fellow of the IEEE, AAAS, and AIAA (Asia-Pacific Artificial Intelligence Association) and a past member of the IFIP Committee on System Modeling and Optimization. He was awarded the 2020 Research Impact Award by IEEE Technical Committee on Cloud Computing. He was also awarded the 2022 Distinguished Alumnus Award from the Indian Institute of Technology, Kanpur. His research is currently funded by NIH, NSF, USDOT, DOE and FDOT.
\end{IEEEbiography}

\vspace{-1.25cm}

\begin{IEEEbiography}[{\includegraphics[width=0.8in,height=1in,clip]{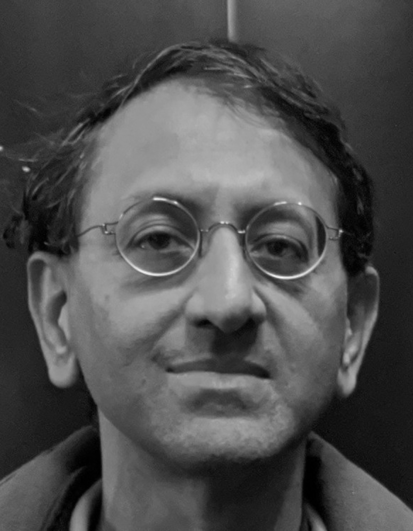}}]%
{Anand Rangarajan} is  a Professor in the Department of Computer and Information Science and
Engineering, University of Florida, Gainesville, FL,
USA. His research interests are computer vision,
machine learning, medical and hyperspectral imaging and the science of consciousness.
\end{IEEEbiography}


\end{document}